\PassOptionsToPackage{table}{xcolor}
\documentclass[10pt,twocolumn,letterpaper]{article}

\usepackage[pagenumbers]{iccv}              

\usepackage{bbding}
\usepackage{url}
\usepackage{tcolorbox}
\usepackage{multirow}
\usepackage{booktabs}
\usepackage{makecell}
\usepackage{float}
\usepackage[T1]{fontenc} 
\usepackage{arydshln}
\usepackage{tcolorbox}
\usepackage{colortbl}

\definecolor{iccvblue}{rgb}{0.21,0.49,0.74}
\usepackage[pagebackref,breaklinks,colorlinks,allcolors=iccvblue]{hyperref}

\def\paperID{10888} 
\def\confName{ICCV}
\def\confYear{2025}

\newcommand\blfootnote[1]{%
  \begingroup
  \renewcommand\thefootnote{}\footnote{#1}%
  \addtocounter{footnote}{-1}%
  \endgroup
}

\title{VideoOrion: Tokenizing Object Dynamics in Videos}

\author{
Yicheng Feng\textsuperscript{\rm 1}$^\dagger$ \ \  Yijiang Li\textsuperscript{\rm 3,2}$^\dagger$ \ \  Wanpeng Zhang\textsuperscript{\rm 1} \ \  Hao Luo\textsuperscript{\rm 1} \ \ \\ Zihao Yue\textsuperscript{\rm 4} \ \ Sipeng Zheng\textsuperscript{\rm 2} \ \  Zongqing Lu\textsuperscript{\rm 1,2}$^\ddagger$\\ \\
\textsuperscript{\rm 1}School of Computer Science, Peking University, 
\textsuperscript{\rm 2}Beijing Academy of Artificial Intelligence  \\
\textsuperscript{\rm 3}University of California, San Diego, 
\textsuperscript{\rm 4}Renmin University of China
}

\begin{document}
\twocolumn[{%
\maketitle
\vspace{-0.3in}
\begin{figure}[H]
\hsize=\textwidth %
\centering
\includegraphics[width=0.9\textwidth]{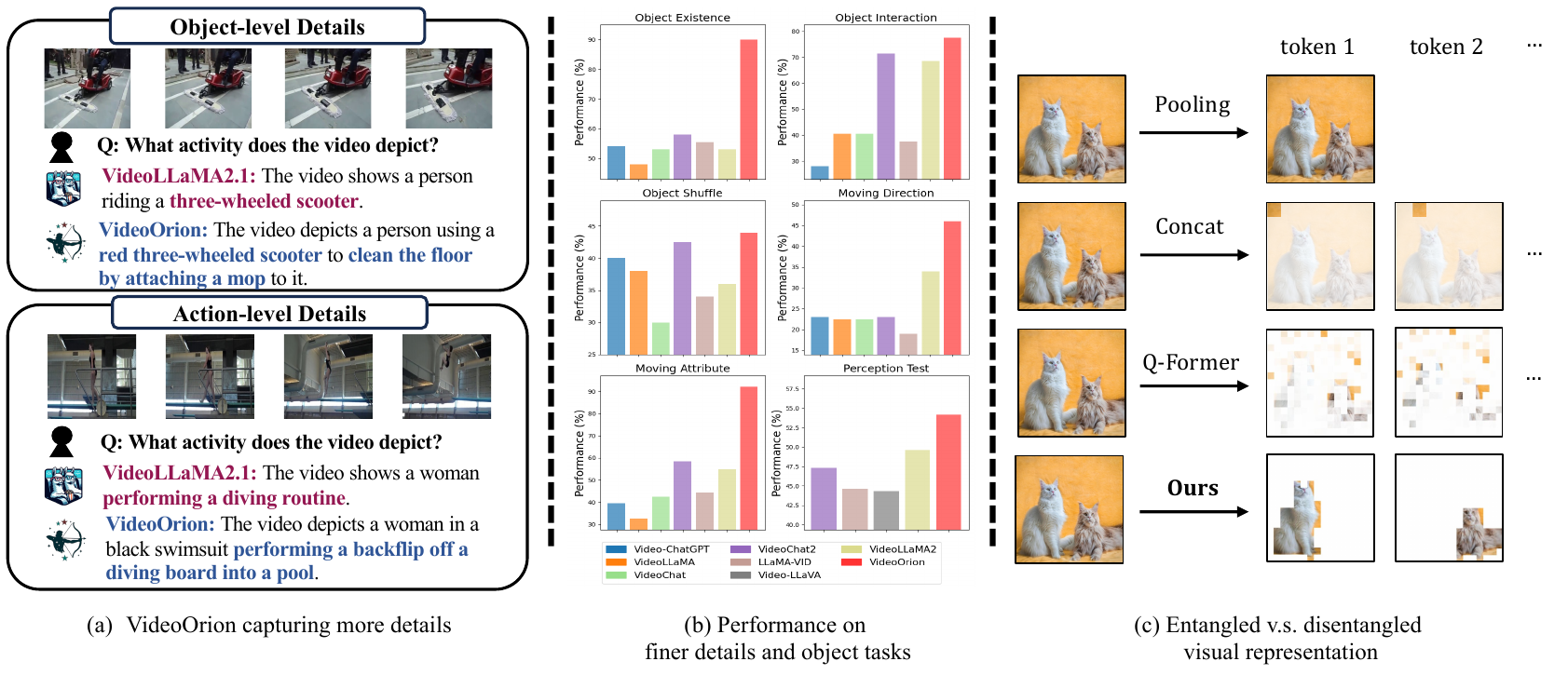}
    \vspace{-2mm}
    \caption{With explicit modeling of object dynamics, \textit{VideoOrion} can (a) grasp finer details (b) with understanding on object-related fine-grained details. (c) Comparison with prior encoding including: (1) spatial pooling the whole frame into a single token; (2) concatenating adjacent patch tokens into a single token; (3) Q-Former aggregates patch tokens with learnable queries. (4) \textit{VideoOrion} with \textit{object tokens} providing disentangled semantics.}
    \label{fig:intro}
\end{figure}
}]
\blfootnote{$\dagger$ Equal contribution.}
\blfootnote{$\ddagger$ Correspondence to <zongqing.lu@pku.edu.cn>}

\maketitle
\begin{abstract}
We present VideoOrion, a Video Large Language Model (Video-LLM) that explicitly captures the key semantic information in videos—the spatial-temporal dynamics of objects throughout the videos.
VideoOrion employs expert vision models to extract object dynamics through a detect-segment-track pipeline, encoding them into a set of \textit{object tokens} by aggregating spatial-temporal object features. 
Our method addresses the persistent challenge in Video-LLMs of efficiently compressing high-dimensional video data into semantic tokens that are comprehensible to LLMs. Compared to prior methods which resort to 
downsampling the original video or aggregating visual tokens using resamplers, leading to information loss and entangled semantics, VideoOrion not only offers a more natural and efficient way to derive compact, disentangled semantic representations but also enables explicit object modeling of video content with minimal computational cost. Moreover, the introduced object tokens naturally allow VideoOrion to accomplish video-based referring tasks. Experimental results show that VideoOrion can learn to make good use of the object tokens, and achieves competitive results on both general video question answering and video-based referring benchmarks.
\end{abstract}    
\section{Introduction}
\label{sec:intro}
The remarkable performance of Large Language Models (LLMs) \citep{achiam2023gpt,dubey2024llama,team2023gemini} has spurred interest in extending these models' capabilities beyond text, which catalyzes the development of multi-modal large language models (MLLMs)  \citep{liu2024visual,bai2023qwen,chen2023minigpt,zhang2023video,shu2023audio}. 
By processing and integrating diverse modalities through tokenization and alignment with text tokens, MLLMs enable a broader range of real-world applications.
However, a significant challenge lies in efficiently encoding the information from multi-modal inputs into a limited number of tokens, particularly for Video-LLMs, as videos encapsulate much more complex and detailed information than other input modalities, such as an image.

To compress high-dimensional visual information into a more compact representation, existing studies commonly employ downsampling or pooling techniques before tokenization \citep{maaz2023video,jin2024video} and integrate various aggregation modules to reduce the number of visual tokens \citep{li2023videochat, cheng2024videollama, alayrac2022flamingo}, thereby mitigating computational costs. Several limitations arise. First, due to computational constraints, Video-LLMs usually only sample a small fraction of the frames in the video for training and inference (e.g., sample 8 or 16 frames out of thousands of frames for a video of about three minutes). 
Despite being more efficient, they inevitably incur information loss, particularly in the fine-grained dynamics of objects and interactions between scenes. The discretized frames also fail to provide sufficient contextual information for Video-LLMs to effectively model long-range temporal dependencies within videos. We hypothesize that this limitation is a key factor preventing current video-LLMs from achieving a more detailed understanding of video content, restricting their ability to capture intricate nuances beyond a general overview. Furthermore, existing methods encode video tokens by processing image patches through a vision encoder, often overlooking the explicit semantics embedded within these visual tokens. This oversight can lead to semantically entangled representations \citep{xu2024slot}. In contrast, text tokens inherently carry clear and well-defined semantics, creating a significant disparity that complicates the alignment of ambiguous visual tokens with semantically precise text tokens in the LLM.

In this paper, we present \textit{VideoOrion}, a Video-LLM with a novel vision encoding method that explicitly captures the key semantic information in videos. Drawing inspiration from the way humans naturally identify object semantics from visual observations before integrating contextual information into cognitive processes, we argue that a tokenizer for the visual modality should provide semantically rich tokens, akin to tokenizers in text processing, to enhance a model’s understanding of visual content. A key aspect of visual semantics lies in the object dynamics, including their appearance, interactions, and temporal variations.

To effectively capture these dynamics, \textit{VideoOrion} introduces a detect-segment-track pipeline that utilizes expert models to extract objects and their evolving characteristics across a sequence of frames. This information is then fused into a set of \textit{object tokens}, representing the objects and their spatial-temporal dynamics throughout the video. Beyond object dynamics, contextual visual information is also essential, serving as complementary to object tokens for a more comprehensive understanding of the video. To address this, we propose to also supplement the object tokens with a set of \textit{context tokens} produced by a Video Projector. By incorporating both encoding branches, \textit{VideoOrion} effectively captures overarching contextual elements (e.g., static scene information) while also preserving fine-grained details about specific objects or instances through object tokens. This disentangled object representation enables the subsequent LLM to more accurately model spatial and temporal interactions between objects, ultimately improving video comprehension.


 Through extensive experiments and visualization, \textit{VideoOrion} demonstrates superior performance in video understanding tasks across multiple benchmarks. 
 Moreover, the proposed object tokens naturally facilitate video-based referring, i.e., visual question-answering involving a specific object or instance in the videos (provided in the first frame)\citep{qiu2024artemis,yu2025merlin,wang2024elysium}. Notably, \textit{VideoOrion} exhibits remarkable capabilities on video-based referring tasks, achieving substantial gains compared to previous methods. Our main contributions can be summarized as follows:

\begin{itemize}
    \item We present \textit{VideoOrion}, featuring a novel object branch that encodes the spatial-temporal dynamics of objects and instances in the videos through a set of \textit{object tokens}.
    \item To effectively capture object dynamics, we propose a detect-segment-track pipeline that leverages knowledge from expert vision models to extract object masks across frames, thereby explicitly generating disentangled objects representations.
    \item  We conduct extensive experiments and ablation studies on multiple benchmarks, showcasing consistent improvements with object tokens and achieving competitive results on general VQA and video-based referring tasks.
\end{itemize}

\begin{figure*}[h]
     \centering
     \includegraphics[width=0.85\textwidth]{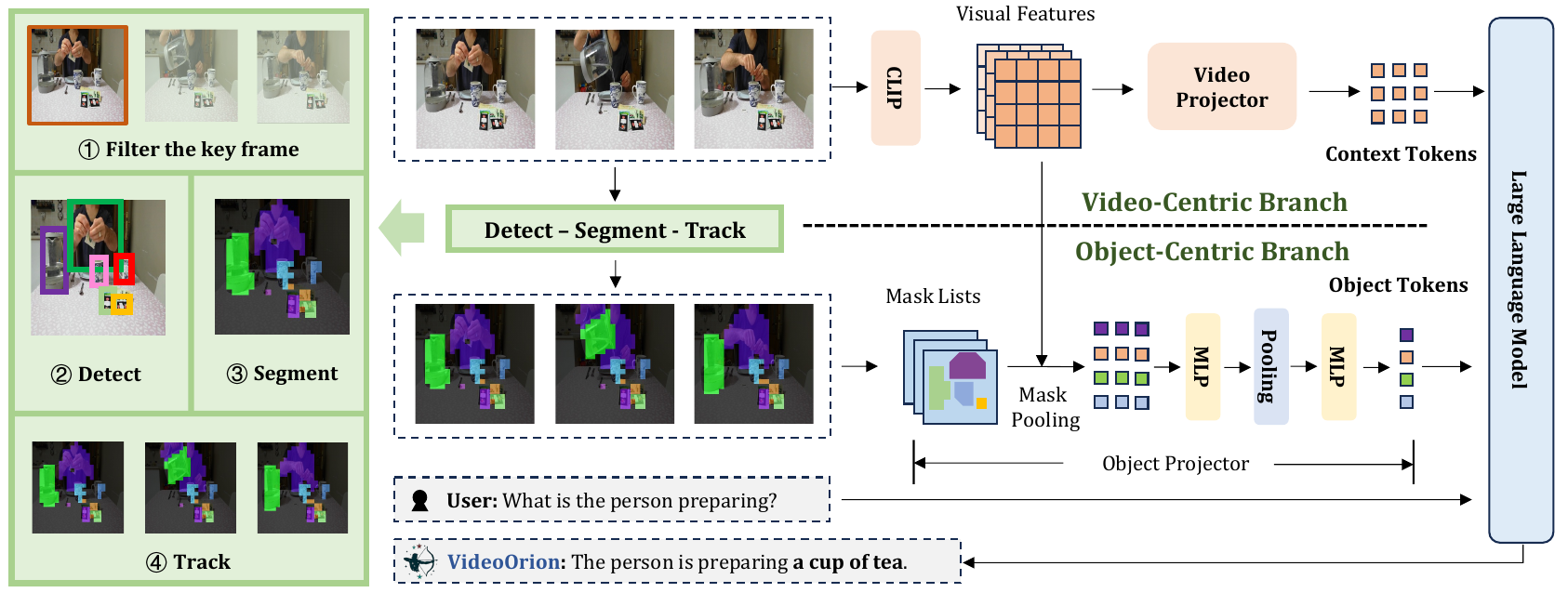}
     \vspace{-2mm}
     \caption{The overall architecture of \textit{VideoOrion}. Two branches are employed to encode the video content into tokens: the Video-Centric Branch encodes the general information with \textit{context tokens}, while the Object-Centric Branch encodes the dynamics of objects through the detect-segment-track pipelines in the video into a set of \textit{object tokens}. All these tokens are fed together to the LLM for integrating information from both branches and generating responses to the text inputs.}
     \label{fig:overview}
     \vspace{-2mm}
\end{figure*}
\section{Related Work}
\textbf{Video-LLMs.}
Most existing Video-LLMs extend the framework of image-based MLLMs by introducing various adapters to handle the large number of visual tokens generated by image-level pretrained visual encoders such as CLIP~\citep{radford2021learning}. Prominent models include Video-ChatGPT\citep{maaz2023video}, which employs spatial-temporal pooling; VideoChat\citep{li2023videochat}, VideoChat2\citep{li2024mvbench}, and VideoLLaMA\citep{zhang2023video}, which leverage Q-Former~\citep{li2023blip,zhu2023minigpt}; MiniGPT4-Video\citep{ataallah2024minigpt4}, which concatenates adjacent visual tokens and applies linear mapping; VideoLLaMA2\citep{cheng2024videollama}, which incorporates 3D convolution blocks; and LLaMA-VID~\citep{li2023llama}, which utilizes cross-attention between visual and textual embeddings. While these approaches have achieved promising results, they predominantly emphasize adapting visual features for integration into large language models (LLMs), often neglecting explicit semantic representation within visual tokens. Consequently, the task of interpreting token semantics is largely delegated to the LLMs themselves.

There are also studies emphasizing the semantic aggregation of visual tokens. For instance, Chat-univi~\citep{jin2024chat} aggregates visual tokens into dynamic clusters using DPC-KNN~\citep{dpc-knn}, a clustering algorithm aimed at reducing redundancy by merging visually similar tokens. Unlike Chat-univi, \textit{VideoOrion} aggregates visual tokens explicitly in an object-centric manner, thus ensuring clearer semantic representation. Similarly, Video-LaVIT~\citep{jin2024video} employs the MPEG-4 compression algorithm to distill video information into key frames and motion features, subsequently encoding these using VQ-VAE. While our approach also segments videos via key frames, we prioritize the semantic representation of objects rather than general motion information. The closest related study is Slot-VLM~\citep{xu2024slot}, which shares conceptual similarities with ours by employing slot attention mechanisms to cluster tokens into object-centric and event-centric representations. Our methodology substantially diverges by deploying specialized vision models to precisely detect and extract object representations and incorporating temporal fusion techniques to capture the dynamics of objects across frames. This ensures that our resulting object tokens encapsulate richer, temporally integrated semantic information. Artemis~\citep{qiu2024artemis} also incorporates ROI tracking to derive target-specific features from visual tokens but predominantly targets video-based referential understanding tasks with exclusive emphasis on individual objects. In contrast, our research aims to enhance the object-centric comprehension of Video-LLMs by systematically aggregating and encoding semantic tokens for all identified objects within the observed video sequences.


\textbf{Vision Foundation Models.}
Humans naturally incorporate visual information into reasoning by semantically segmenting and extracting meaningful visual content, particularly when identifying distinct objects or entities. However, existing Multimodal Large Language Models (MLLMs) often neglect this crucial semantic abstraction. Notably, this semantic-level processing aligns closely with several fundamental computer vision tasks, including object detection \citep{carion2020end,zhang2022dino,oquab2023dinov2,yao2022detclip}, instance segmentation \citep{he2017mask,cheng2022masked}, and video object segmentation \citep{perazzi2017learning,wang2019fast,yang2021associating}. 

To bridge this gap, we propose to leverage specialized vision foundation models to more effectively encode object dynamics, rather than exclusively relying on video-text paired approaches such as CLIP \cite{radford2021learning}.  For instance, GroundingDINO \citep{liu2023grounding}  introduces open-vocabulary object detection capabilities by marrying DINO \citep{zhang2022dino, oquab2023dinov2} with grounding abilities by contrastive training on object region-text pairs \cite{yao2022detclip, carion2020end}.  Moving beyond text-based prompts, another line of research proposes SAM \citep{cheng2022xmem, ravi2024sam} with the ability to interactive segment anything with points, scribbles, and bounding boxes. Further, beyond understanding a single image, multi-object tracking \citep{zhang2022bytetrack,mancusi2023trackflow,qin2024towards} and object segmentation \citep{perazzi2017learning,wang2019fast,yang2021associating}  in video contexts have been extensively studied, resulting in the emergence of several specialized vision foundation models \citep{cheng2022xmem, ravi2024sam, zhu2023tracking,meinhardt2021trackformer}. 

\section{Methodology}
\label{method}

\begin{table*}[h]
\centering
\caption{\small Data mixture of \textit{VideoOrion} training. For video-centric branch, we directly utilized open-source base models.}
 \label{tab:dataset}
 \vspace{-2mm}
 \resizebox{0.8\textwidth}{!}{
 \begin{tabular}{ccccl}
 \toprule
 Stage & Modality & \# Samples & Object Tokens & Source  \\
 \midrule
 Video pretraining & \multicolumn{4}{c}{VideoLLaMA2 Base Model} \\
\hdashline
 Object pretraining & video-text & 700K & $\surd$ & InternVid-10M, OpenVid-1M \\
\hdashline
 \multirow{4}{*}{MM instruction tuning} & video-text & 2.7M & $\surd$ & Video-ChatGPT, VideoChat2, LLaVA-Video \\
 & video-text & 1M & $\surd$ & Ego4D, EgoExo4D \\
 & image-text & 625K & $\surd$ & LLaVA \\
 & text-only  & 40K & & Video-LLaVA \\
 \bottomrule
 \end{tabular}
 }
\vspace{-2mm}
\end{table*}

Building upon the discussion presented in Section~\ref{sec:intro}, we introduce \textit{VideoOrion}, an instantiation of an object-centric representation characterized by enriched semantics, increased compactness, and greater efficiency—achieving comprehensive information encoding with fewer tokens. It is important to emphasize that one of our primary contributions lies in this conceptual idea of object-centric representation, with \textit{VideoOrion} serving merely as a specific realization of this broader concept, detailed further below. \textit{VideoOrion} demonstrates the effectiveness of utilizing a more disentangled and compact representation of object dynamics, which enhances and complements general video features, thereby significantly improving video modeling and comprehension.
\subsection{Overall Architecture}
\textit{VideoOrion} features on an Object-Centric Branch complemented by a Video-Centric Branch, designed for comprehensive tokenization of both general video context and specific object dynamics, as depicted in Figure~\ref{fig:overview}.
These two branches function independently, with the downstream large language model (LLM) tasked with integrating semantic information derived from each branch. This decoupled design, naturally enables \textit{VideoOrion} with high flexibility, readily accommodating various architectural configurations, i.e. different video projectors \citep{maaz2023video,li2023blip,zhang2023video, cheng2024videollama, liu2023visual} and LLMs \citep{jiang2023mistral,yang2024qwen2}.

\subsection{Video-Centric Branch}
Given a video $V_i\in \mathbb{R}^{T\times H\times W\times C}$, where $T$ denotes the number of frames. The Video-Centric Branc first utilizes a vision encoder, e.g. CLIP to extract the features before employing a Video Projector to project the high-dimensional per-frame features into a set of \textit{context tokens}. This procedure usually incurs large computation due to the large number of \textit{context tokens} (e.g. 576 per frame) which limits the number of frames to be encoded. Consequently, these tokens can only encode the general and static information, such as the background and scene. \textit{VideoOrion} supports diverse Video Projector designs; here, we employ the STC Connector introduced by VideoLLaMA2 \citep{cheng2024videollama} for its superior preservation of spatial-temporal details, enhancing the association between visual and object tokens. Specifically,  we first sample a fixed number of frames from the video, and use a vision encoder such as CLIP (ViT-L/14) \citep{radford2021learning} to encode each frame, resulting in the embedding $x_i \in \mathbb{R}^{t_v\times h\times w\times D}$, where $t_v$ is the number of sampled frames, and $h=H/p, w=W/p$ are the resolution of visual embeddings, $p=14$ for ViT-L/14. Then, we use STC Connector, which is composed of two RegStage blocks \citep{radosavovic2020designing} and a 3D convolution block, followed by an MLP projector, to transform $x_i$ into context tokens $v_i \in \mathbb{R}^{N_v\times d}$, where $N_v$ is the number of context tokens and $d$ is dimension of the LLM decoder's token embedding space.

\subsection{Object-Centric Branch}
\label{method:obj-branch}
In the Object-Centric Branch, our objective is to extract a concise set of \textit{object tokens} from the entire video, each encapsulating both semantic and spatial-temporal information related to individual objects or instances and their dynamics throughout the video. This process begins with object proposal generation, where instances of interest are identified, followed by dynamic tracking to establish their trajectories across frames. Finally, an Object Projector aggregates the encoded features of these dynamics and trajectories to derive compact \textit{object tokens}.

\noindent \textbf{Detect-segment-track pipeline.} 
Identifying and tracking the dynamics of objects in videos present significant challenges. A fundamental question is: How can we efficiently represent an object along with its temporal dynamics? To achieve this, an ideal representation should capture both the spatial location and fine-grained details of objects throughout a video's progression. Using masks to refer to objects' representations naturally emerge as a superior choice as masks can not only precise spatial information, but also encapsulate precise object contours, free-form shapes, and other fine-grained details (compared to coarse-grained bounding boxes), which have been widely adopted in referring and grounding methods \citep{you2023ferret, ravi2024sam} for object representation. By associating a sequence of masks with an object across multiple frames, we can effectively characterize its motion and evolution within the video.

To identify objects of interest, we employ GroundingDINO \citep{liu2023grounding} in its generic mode during the \textit{detect} stage, as it offers both efficiency and strong performance in generating object region proposals in specific frames. In the subsequent \textit{segment} stage, we leverage SAM \citep{kirillov2023segment} to refine these bounding box proposals into precise object masks. However, due to the dynamic nature of objects and scene changes, objects may enter or exit the frame over time. A naive approach that tracks only the objects present in the first frame risks substantial information loss. A possible strategy is to uniformly sample multiple frames and segment the video into clips for object identification and tracking. However, this approach may lead to inconsistencies, where the same object is misidentified across different clips. Additionally, it may fail to capture newly appearing objects accurately. Another alternative is to use tracking algorithms capable of detecting new objects automatically; however, these methods are typically limited to closed-set object categories, rendering them unsuitable for general video analysis. To address these challenges, we propose a novel segmentation method that dynamically partitions the video based on variations in object presence across frames. Specifically, we sample a sequence of frames, ${f_1, f_2, ..., f_n}$, and apply RAM++ \citep{huang2023open}, an open-world image recognition model, to annotate each frame with a set of object-related tags. We then utilize the NLTK toolkit \citep{bird2009natural} to filter these tags, extracting concrete nouns and their synonyms to construct a tag set ${l_1, l_2, ..., l_n}$, which serves as an indicator of object presence in each frame. To refine this process, we introduce two thresholds, $\theta_a$ and $\theta_b$. The first threshold, $\theta_a$, is used to filter out frames with a low number of object tags, as such frames are often noisy: $\mathcal{F}_1 = \{ f_u \mid u \in \{1, \ldots, n\}, \, |l_u| > \theta_{a} \}.$ The first frame in $\mathcal{F}_1$ is designated as the initial key frame. Subsequently, for each subsequent frame, we compute the overlap between its tag set and that of the most recent key frame. If the overlap falls below the threshold $\theta_b$, i.e., $|l_u \cap l_{key}| < \theta_{b}$. we consider this a significant object transition and designate the frame as a new key frame. The values of $\theta_a$ and $\theta_b$ are empirically determined based on the performance characteristics of RAM++, with our experiments using $\theta_a = 3$ and $\theta_b = 2$.

Through this process, we obtain a sequence of key frames that segment the video into distinct clips based on object transitions. For each key frame, we apply the object proposal and segmentation steps to generate object masks.

In the tracking stage, we sample $t_o$ frames from the video and partition them into clips based on the key frames. Notably, $t_o$ is typically much larger than $t_v$ since it does not affect the number of tokens, thereby keeping computational costs manageable. In our experiments, we set $t_v = 8$ and use $t_o = 64$ for short videos, increasing to $t_o = 128$ for videos longer than one minute. A key advantage of this approach is that object tokens retain richer temporal information compared to context tokens. Finally, we utilize XMem \citep{cheng2022xmem}, a multi-object tracking algorithm, to track all object masks across the entire video, starting from each key frame. This results in a set of object mask lists: $\mathcal{M} = \{M_1, M_2, ..., M_{N_{oi}}\}$ where $N_{oi}$ represents the total number of identified objects in video $V_i$, and $M_j$ corresponds to the mask sequence for object $j$ across frames.

\noindent \textbf{Object Projector.} We introduce Object Projector to translate the list of object masks into \textit{object tokens}. Each object \( j \), appearing in \( k \) frames, has an associated mask list \( M_j = \{m_{j}^{f_{j1}}, m_{j}^{f_{j2}}, \dots, m_{j}^{f_{jk}}\} \). 
Mask pooling is performed on each mask to fuse the features from the corresponding frames \( f_{j1}, \dots, f_{jk} \). Subsequently, temporal pooling, followed by an MLP projector, integrates these spatially pooled features into a unified representation.
Consequently, for each \( V_i \), we obtain \( N_{oi} \) compact \textit{object tokens} \( o_j \in \mathbb{R}^d \), where \( j \in \{1, \ldots, N_{oi}\} \). These object tokens encapsulate both spatial and temporal dynamics effectively.

\subsection{Training}

\textbf{Instruction Template.} To integrate context tokens and object tokens with text tokens for input into the LLM, we define the following input template:

\begin{tcolorbox}[boxrule=0.2mm]
\textbf{User:} \texttt{<$v$>} Here is a list of objects and instances in the video: \texttt{<$o_1$>},\texttt{<$o_2$>},...,\texttt{<$o_N$>} \texttt{<Instruction>} \\
\textbf{Assistant:} \texttt{<Answer>}
\end{tcolorbox}

\noindent where \texttt{<$v$>} represents the context tokens of the video sample, and \texttt{<$o_1$>}, \texttt{<$o_2$>}, ..., \texttt{<$o_N$>} denote the object tokens.
Following the two-branch design of \textit{VideoOrion}, the training procedure is divided into three stages: 1) Video-Centric Branch pretraining; 2) Object-Centric Branch pretraining; 3) Multi-modal instruction tuning.

\noindent \textbf{Video-Centric Branch pretraining.} 
At this stage, only the Video Projector is optimized using video-text and image-text pairs for vision-language alignment, without incorporating object tokens. To ensure simplicity and computational efficiency, we directly utilize the open-source base model of the STC-Connector from VideoLLaMA2\citep{cheng2024videollama}, which has been pretrained on 12.2M vision-language data for vision-language alignment \citep{cheng2024videollama}.

\noindent \textbf{Object-Centric Branch pretraining.} At this stage, the pretrained Video Projector remains frozen, and only the Object Projector is optimized using video-text pairs. We construct a training dataset by sourcing captioned videos from InternVid-10M \citep{wang2023internvid} and OpenVid-1M \citep{nan2024openvid}. To refine the dataset, we first filter InternVid-10M using aesthetic scores\citep{schuhmann2022laion} and UMT-SIM scores\citep{li2023unmasked}, which assess video quality and video-caption similarity, respectively, yielding a 1.4M-sample subset. This subset is then combined with OpenVid-1M, and the resulting dataset undergoes further filtering based on noun-phrase concept balance, following \citep{liu2024visual}. Ultimately, we obtain a pretraining dataset comprising 700K video-text pairs. To enhance training efficiency, we preprocess this dataset using a detect-segment-track pipeline to generate object mask lists.

\noindent \textbf{Multi-modal instruction tuning.} In the final stage, we only freeze the vision encoder, and optimize the Video Projector, the Object Projector, and the LLM backbone together, with multi-modal (MM) instruction tuning datasets.
We adopt the training data from Video-LLaVA\citep{lin2023video}, VideoChat2\citep{li2024mvbench} and LLaVA-Video\citep{zhang2024llavanextvideo}. We also incorporate an ego-centric video question-answering dataset built with videos and annotations from Ego4D\citep{grauman2022ego4d} and EgoExo4D\citep{grauman2024ego}. We sampled 1M ego data samples to make up to the 4M dataset size.
We segment egocentric videos into shorter clips and generate question-answer pairs based on the original human annotations as well as additional annotated visual information. These question-answer pairs encompass multi-level knowledge, ranging from low-level visual basic facts to high-level behavior-centric understanding, and feature diverse types of questions, spanning from descriptive to deductive in nature.
We preprocess the video-text samples and image-text samples with the detect-segment-track pipeline to obtain the object mask lists here. For images, we treat them as single-frame videos, and the tracking model is not used in the pipeline.

We use auto-regressive cross-entropy loss in all three stages. The data mixture is summarized in Table~\ref{tab:dataset}.

\section{Experiments}
\subsection{Experimental Setup}

We present two variants of \textit{VideoOrion}: \textit{VideoOrion}, which uses CLIP (ViT-L/14) \citep{radford2021learning} as the vision encoder and Mistral-Instruct-7B \citep{jiang2023mistral} as the LLM backbone; and \textit{VideoOrion$+$}, which uses SigLIP (so400m-patch14-384) \citep{zhai2023sigmoid} as the vision encoder and Qwen2-7B \citep{yang2024qwen2} as the LLM backbone.
We sample $t_v=8$ frames from each video for the Video-Centric Branch for \textit{VideoOrion}, and set $t_v=16$ for \textit{VideoOrion$+$}. We sample $t_o=64$ frames from short videos and $t_o=128$ frames from videos longer than 1 minute for the Object-Centric Branch. We set $\theta_a=3$ and $\theta_b=2$ when extracting key frames to split videos with RAM++, which we find have the best performance, and the number of sampled frames for tagging is set to $n=16$ for Object-Centric Branch pretraining and $n=64$ for MM instruction tuning. We resize and crop each frame to $336\times 336$ for CLIP \citep{radford2021learning} and $384\times 384$ for SigLIP \citep{zhai2023sigmoid}, following their implementations. We use a learning rate of $1e-4$ and $5e-6$ for the two stages, with a warm-up ratio of $0.03$. The global batch size for Object-Centric Branch pertaining is $256$ and $128$ for the MM instruction tuning stage. In both stages, we train for only one epoch, with $8\times$ A$800$ GPUs, 
for \textit{VideoOrion} and $64\times$ A$800$ GPUs for \textit{VideoOrion$+$}.

\subsection{Video Question Answering}
\textbf{Evaluation benchmarks.}
We comprehensively evaluate the video understanding capabilities of \textit{VideoOrion} and \textit{VideoOrion$+$} on four multi-choice video question answering (MC-VQA) datasets including MVBench \citep{li2024mvbench}, EgoSchema \citep{mangalam2023egoschema}, Perception-Test \citep{patraucean2024perception} and VideoMME \citep{fu2024video}. Accuracies are reported for each of the benchmarks. We also test on an open-ended video question answering (OE-VQA) benchmark ActivityNet-QA \citep{yu2019activitynet}, where ChatGPT3.5 is employed to evaluate the answer following \citet{maaz2023video}, by two metrics: a yes/no indicator that signifies whether the predicted answer matches the correct answer, and a score ranging from 0 to 5 that reflects the degree of alignment between the model output and the correct answer. We report both the accuracy and the average score.

\begin{table*}[h]
\centering
 \caption{\small Performance comparison with the state-of-the-art Video-LLMs. All models except ShareGPT4Video use a 7B LLM backbone.}
 \label{tab:main-result}
 \vspace{-2mm}
 \resizebox{0.8\textwidth}{!}{
 \begin{tabular}{lcccccc}
 \toprule
 Model & \makecell{Frame\\Number} & \makecell{MVBench \\ Acc.} & \makecell{Egoschema \\ Acc.} & \makecell{Perception-Test \\ Acc.} & \makecell{Video-MME \\ w/o / w subs} & \makecell{ActivityNet-QA \\ Acc. / Score} \\
 \midrule
 LLaMA-VID\citep{li2023llama} & 1fps & 41.9 & 38.5 & 44.6 & 25.9/~ - ~ & 47.4/3.3 \\
 TimeChat\citep{ren2024timechat} & - & 38.5 & 33.0 & - & - & -\\ 
 Chat-UniVi\citep{jin2024chat} & 64 & - & - & - & 40.6/45.9 & 46.1/3.3 \\
 LLaVA-NeXT-Video\citep{zhang2024llavanextvideo} & 32 & 46.5 & 43.9 & 48.8 & 33.7/~ - ~ & \underline{53.5}/3.2 \\
 ShareGPT4Video-8B\citep{chen2024sharegpt4video} & 16 & 51.2 & - & - & 39.9/43.6 & - \\
 VideoChat2\citep{li2024mvbench} & 16 & \underline{60.4} & \underline{54.4} & 47.3 & 39.5/43.8 & 49.1/3.3 \\
 Video-LLaVA\citep{lin2023video} & 8 & 41.0 & 38.4 & 44.3 & 39.9/41.6 & 45.3/3.3 \\
 VideoLLaMA\citep{zhang2023video} & 8 & 34.1 & - & - & - & 12.4/1.1 \\
 VideoLLaMA2\citep{cheng2024videollama} & 8 & 53.4 & 50.5 & \underline{49.6} & \underline{45.1}/\underline{46.6} & 49.9/\underline{3.3} \\
 \rowcolor{gray!20} VideoOrion & 8 & \textbf{63.5} & \textbf{65.1} & \textbf{65.2} & \textbf{54.6}/\textbf{55.3} & \textbf{57.7}/\textbf{3.7} \\
 \midrule
 \multicolumn{7}{l}{\textcolor{gray}{\textit{Models with Qwen-2-7B LLM backbone}}} \\ 
 LongVA\citep{zhang2024long} & 64 & - & - & - & 52.4/~ - ~ & ~ - ~/2.8 \\
LLaVA-OneVision\citep{li2024llava} & 32 & 56.7 & \underline{60.1} & \underline{57.1} & \underline{58.2}/\textbf{61.5} & \underline{56.6}/~ - ~ \\
 VideoLLaMA2.1\citep{cheng2024videollama} & 16 & \underline{57.3} & 53.1 & 54.9 & 54.9/\underline{56.4} & 53.0/\underline{3.4} \\
  \rowcolor{gray!20} VideoOrion$+$ & 16 & \textbf{67.4} & \textbf{65.0} & \textbf{65.9} & \textbf{58.9} / \textbf{61.5} & \textbf{60.3}/\textbf{3.7} \\
 \bottomrule
 \end{tabular}
 }
 \vspace{-2mm}
\end{table*}

\noindent \textbf{Zero-shot performance.} We compare the performance of \textit{VideoOrion} and \textit{VideoOrion$+$} with prior state-of-the-art Video-LLMs in Table~\ref{tab:main-result}. 
\textit{VideoOrion} achieves competitive performance consistently surpassing the second-best methods by a large margin.
Notably, compared to the baselines VideoLLaMA2 and VideoLLaMA2.1 which has the same Video-Centric Branch, \textit{VideoOrion} and \textit{VideoOrion$+$} consistently outperforms them by $10.1\%$, $14.6\%$, $15.6\%$, $8.7\%$, $7.8\%$ and $10.1\%$, $11.9\%$, $11.0\%$, $5.1\%$, $7.3\%$ on MVBench, EgoSchema, Perception-Test, VideoMME and ActivityNet-QA respectively.
This result demonstrates the effectiveness of Object-Centric Branch for general video understanding, serving as a proof of concept that explicit disentangled object presentation can efficiently and effective encode the rich information in videos for LLMs to comprehend.

\subsection{Video-based Referring}

With the proposed Object-Centric Branch, \textit{VideoOrion} inherently supports video referring—a capability often absent in conventional Video-LLMs. To enable video referring, we simply structure the input prompt template as follows:
\vspace{-2mm}
\begin{tcolorbox}[boxrule=0.2mm]
\textbf{User:} \texttt{<$v$>} What is \texttt{<$o$>} doing in this video? \\
\textbf{Assistant:} \texttt{<Answer>}
\end{tcolorbox}
\vspace{-1mm}

\noindent where we insert the object token corresponding to the referring target in the video with \texttt{<$o$>} in the instruction. We evaluate performance on the VideoRef45K benchmark \citep{qiu2024artemis}, which comprises video question-answer data with box-level prompts in the first frame to specify the referring target. To encode object tokens, we apply SAM to the bounding box prompts to extract target masks.

We train two variants of models with a subset of VideoRef45K including data from VID-Sentence~\citep{chen2019weakly} (8K), HC-STVG~\citep{tang2021human} (10K) and LaSOT~\citep{fan2019lasot} (8K). For \textit{VideoOrion-Ref}, we integrate this data into the MM instruction tuning stage with instruction following data from VideoLLaVA to train \textit{VideoOrion}. For \textit{VideoOrion-Ref-FT} and \textit{VideoOrion-Ref-FT$+$}, we finetune the \textit{VideoOrion} and \textit{VideoOrion$+$} with the referring data for 3 epochs, following \citep{qiu2024artemis}. We evaluate the models with metrics BLEU@$4$~\citep{papineni2002bleu}, METEOR~\citep{banerjee2005meteor}, ROUGE~$\_$L\citep{lin2004rouge}, CIDEr~\citep{vedantam2015cider} and SPICE~\citep{anderson2016spice}. We compare the results with Artemis, a video-based referring model, and Merlin~\citep{yu2025merlin}, a multi-frame-based referring model, in Table~\ref{tab:vidref}.

\begin{table}[h]
\centering
 \caption{\small Performance on the video referring task \citet{qiu2024artemis}.}
 \label{tab:vidref}
 \vspace{-2mm}
 \setlength{\tabcolsep}{2pt}
 \resizebox{0.46\textwidth}{!}{
 \begin{tabular}{lccccc}
 \toprule
 Model & BLEU@$4$ & METEOR & ROUGE$\_$L & CIDEr & SPICE \\
 \midrule
 Merlin\citep{yu2025merlin} & 3.3 & 11.3 & 26.0 & 10.5 & 20.1 \\
 Artemis\citep{qiu2024artemis} & 15.5 & 18.0 & 40.8 & 53.2 & 25.4\\
 \hdashline
 VideoOrion-Ref & 17.5 & 19.5 & 43.0 & 69.7 & 28.4 \\
 VideoOrion-Ref-FT & \underline{19.0} & \underline{21.0} & \underline{43.8} & \underline{79.6} & \underline{30.4} \\
 VideoOrion-Ref-FT$+$ & \textbf{19.7} & \textbf{21.5} & \textbf{45.4} & \textbf{90.6} & \textbf{31.4} \\
 \bottomrule
 \end{tabular}
 }
\vspace{-2mm}
\end{table}

We can see that all our models outperform the baselines on all evaluation metrics. Notably, \textit{VideoOrion-Ref} shows good zero-shot performance, with only a small amount of referring data involved in the MM instruction tuning stage. With additional finetuning following \citep{qiu2024artemis}, \textit{VideoOrion-Ref-FT} and \textit{VideoOrion-Ref-FT$+$} achieve significantly better results. This result validates that object tokens effectively encode accurate object semantics, enabling the model to identify the target object. Moreover, our approach equips Video-LLMs with a unified interface for improved general video understanding and referring capabilities.
\subsection{Ablation Study}
The objective of this section is to understand the effectiveness of each component and how they contribute to the improved performance of \textit{VideoOrion}. Due to limited computation, we conduct all the ablation studies with a subset of the data used in \textit{VideoOrion}. For Video-Centric Branch pertaining, we use 702K video-text pairs provided by Valley \citep{luo2023valley} and 558K image-text pairs provided by LLaVA \citep{liu2024visual}. We use our filtered 700k samples for the Object-Branch pertaining, and use 765K samples form Video-LLaVA\citep{lin2023video} for the MM instruction tuning.

\paragraph{Object Tokens.}
To validate the effectiveness of Object-Centric Branch, we compare \textit{VideoOrion} with baseline model VideoLLaMA2, with the same amount of data. Since the baseline model VideoLLaMA2 does not have the Object-Branch pertaining stage, the 700K data is added to the Video-Branch pertaining stage, for a fair comparison. As per Table ~\ref{tab:ablate-obj-branch}, \textit{VideoOrion} with object tokens consistently improves over the baseline on all the benchmarks, showing the effectiveness of explicit object-centric representation.

\begin{table}[h]
\centering
 \caption{\small Ablation study on the Object-Centric Branch.}
 \label{tab:ablate-obj-branch}
 \vspace{-2mm}
 \setlength{\tabcolsep}{2pt}
 \resizebox{0.46\textwidth}{!}{
 \begin{tabular}{lccccc}
 \toprule
 \makecell[l]{Model} & \makecell{MVBench} & \makecell{Egoschema} & \makecell{Perception} & \makecell{VideoMME} & \makecell{ActNet} \\
 \midrule
 video-only& 41.9 & 41.3 & 43.6 & 44.1 & 43.0  \\
 VideoOrion & \textbf{44.2} & \textbf{44.5} & \textbf{46.3} & \textbf{46.1} & \textbf{43.3} \\
 \bottomrule
 \end{tabular}
 }
\vspace{-1mm}
\end{table}

\noindent \textbf{The Object-Centric Branch Pretraining Stage.} We assess the impact of the additional Object-Centric Branch pretraining stage by comparing it to a randomly initialized Object Branch. As shown in Table~\ref{tab:ablate-obj-pretrain}, pretraining consistently improves performance across tasks, except for the Perception Test, highlighting the necessity of pretraining object tokens—similar to standard visual tokens—for effective text alignment.

\begin{table}[h]
\centering
\setlength{\tabcolsep}{4pt}
 \caption{\small Ablation on the Object-Centric Branch pretraining.}
 \label{tab:ablate-obj-pretrain}
 \vspace{-2mm}
 \setlength{\tabcolsep}{2pt}
 \resizebox{0.46\textwidth}{!}{
 \begin{tabular}{cccccc}
 \toprule
 \makecell[l]{Object Pretrain} & \makecell{MVBench} & \makecell{Egoschema} & \makecell{Perception} & \makecell{VideoMME} & \makecell{ActNet} \\
 \midrule
 \XSolidBrush  & 51.2 & 43.5 & \textbf{50.5} & 46.8 & 44.3 \\
 \Checkmark & \textbf{52.5} & \textbf{51.3} & 49.7 & \textbf{47.6} & \textbf{46.3} \\
 \bottomrule
 \end{tabular}
 }
\vspace{-1mm}
\end{table}

\noindent\textbf{Design choices of detect-segment-track pipeline.}
Table~\ref{tab:ablate-pipeline} analyzes the design choices of the detect-segment-track pipeline. By default (Section \ref{method:obj-branch}), we use GroundingDINO (generic mode) for object proposals, RAM++ for adaptive segmenting the video, and XMem for tracking. To evaluate alternatives, we replace GroundingDINO (generic mode) with RAM++ and Mask2Former \cite{cheng2022masked} for object proposal. For segmenting videos, we explore alternatives of segmenting videos uniformly into four parts or using the entire video without any segmentation. For tracking, we substitute Xmem with SAM2 as an alternative. As per Table~\ref{tab:ablate-pipeline}, we show that our detect-segment-track pipeline remains robust across variations. Notably, uniform segmentation slightly underperforms RAM++, offering a trade-off between efficiency and performance. All configurations outperform the video-only baseline (Table~\ref{tab:ablate-obj-branch}), highlighting the strength of our object representation.

\begin{figure*}[h]
    \centering
    \includegraphics[width=0.7\textwidth]{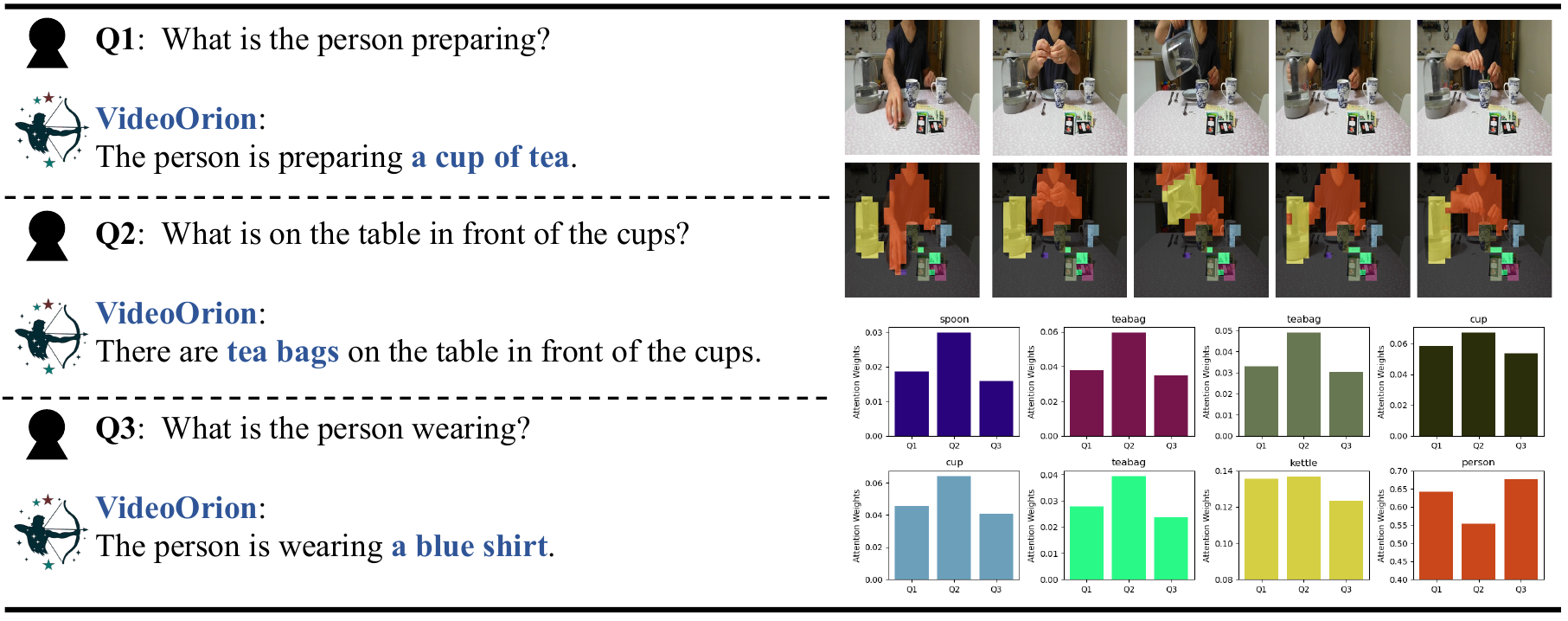}
    \vspace{-2mm}
    \caption{Case studies showing how \textit{VideoOrion} utilizes object tokens to generate responses based on different instructions.}
    \label{fig:case}
    \vspace{-2mm}
\end{figure*}


\begin{table}[h]
\centering
 \caption{\small Ablation study on design choices of the detect-segment-track pipeline. \textit{no-split} refers to not segmenting the video at all; \textit{uniform} refers to uniformly sampling 4 frames as key frames; \textit{M2F} refers to using Mask2Former as object proposer. We color the default choice in grey.}
 \label{tab:ablate-pipeline}
 \vspace{-2mm}
 \setlength{\tabcolsep}{2pt}
 \resizebox{0.35\textwidth}{!}{
 \begin{tabular}{ccccccc}
 \toprule
    Choices  & \makecell{MVB} & \makecell{EGO} & \makecell{PER} & \makecell{V-MME} & \makecell{ACT} & \makecell{Avg.} \\
     
 \midrule

 \midrule
  \midrule
\multicolumn{7}{c}{how to segment the video?} \\
  \midrule
   \midrule
 \rowcolor{gray!20} RAM++ & \textbf{44.2} & \textbf{44.5} & 46.3 & 46.1 & 43.3 & \textbf{44.9} \\
 no-split& 43.8 & 41.2 & 45.3 & \textbf{47.4} & 41.9 & 43.9 \\
 uniform & 43.6 & 43.0 & 45.3 & \textbf{47.4} & \textbf{44.5} & 44.7 \\
  \midrule
   \midrule
\multicolumn{7}{c}{object proposals} \\
  \midrule
   \midrule
   \rowcolor{gray!20} generic & \textbf{44.2} & \textbf{44.5} & 46.3 & 46.1 & 43.3 & \textbf{44.9} \\
 M2F & 43.2 & 41.6 & 43.6 & 46.4 & 42.3 & 43.4 \\
  RAM++ & \textbf{44.2} & 42.9 & 45.5 & 45.2 & 42.9 & 44.1 \\
 \midrule
  \midrule
\multicolumn{7}{c}{tracking model} \\
  \midrule
   \midrule
   \rowcolor{gray!20} Xmem & \textbf{44.2} & \textbf{44.5} & 46.3 & 46.1 & 43.3 & \textbf{44.9} \\
 SAM2 & 44.0 & 41.8 & \textbf{46.7} & 46.3 & 43.3 & 44.4 \\
 \bottomrule
 \end{tabular}
 }
\vspace{-1mm}
\end{table}

\noindent \textbf{Design choices of Object Projector}
\label{exp:arch-ablation}
The Object Projector aggregates object dynamics captured by mask-pooled features. We explore various design choices, including a simple multi-layer perceptron (MLP), a single linear layer, average pooling, and two temporal modeling approaches—attention\citep{vaswani2017attention} and LSTM\citep{hochreiter1997long}. Although we hypothesized that temporal modeling would enhance object tokens, the results in Table~\ref{tab:ablate-arch} unexpectedly reveal that a simple linear layer performs just as effectively. Additionally, we observe that LSTM ranks second on average, with only a minimal gap compared to the MLP.

We further investigate the use of the DINOv2 vision encoder as an alternative to the CLIP encoder for encoding object features. Since DINOv2 is known for its ability to capture objectness, finer details, and low-level features, it could potentially enhance the modeling of object dynamics in videos, albeit at the cost of an additional vision encoder. However, as shown in the last row of Table~\ref{tab:ablate-arch}, its performance falls short of CLIP, likely due to a lack of language-aligned semantics.

\begin{table}
\centering
 \caption{\small Performance of \textit{VideoOrion} with different Object Projectors. We explore mlp (default), attention, linear, lstm layers and plain average pooling.}
 \label{tab:ablate-arch}
 \vspace{-2mm}
 \setlength{\tabcolsep}{2pt}
 \resizebox{0.46\textwidth}{!}{
 \begin{tabular}{lcccccc}
 \toprule
 Architecture & \makecell{MVBench} & \makecell{Egoschema} & \makecell{Perception} & \makecell{VideoMME} & \makecell{ActNet} & Avg. \\
 \midrule
 \rowcolor{gray!20} mlp (ours)  & 48.0 & \textbf{51.5} & 46.9 & \textbf{45.6} & 41.6 & \textbf{46.73} \\
 Attention  & 48.8 & 45.9 & 47.6 & 44.3 & 41.0 & 45.51 \\
 Linear  & 47.3 & 47.2 & 46.1 & 44.1 & \textbf{42.9} & 45.52 \\
 avg pooling  & 47.3 & 47.5 & 46.0 & 45.4 & 41.9 & 45.61 \\
 LSTM  & \textbf{49.5} & 47.0 & \textbf{47.8} & 45.1 & \textbf{42.9} & 46.45 \\
 \hdashline
 DINOv2-L & 44.1 & 41.7 & 45.1 & 45.8 & 41.5 & 43.6\\
 \bottomrule
 \end{tabular}
 }
\vspace{-2mm}
\end{table}

\noindent \textbf{Comparison with Video Encoders.}
Video encoders share a similar function with our Object-Centric Branch, as they not only capture temporal dynamics but also excel at encoding finer details, leveraging emergent objectness within videos and potentially introducing implicit object modeling. Therefore, in Table~\ref{tab:ablate-video-encoder}, we compare \textit{VideoOrion} with baseline models in which the Object-Centric Branch is replaced by two widely used video encoders: VideoMAE \citep{tong2022videomae} and UMT-L \citep{li2023unmasked}. The results demonstrate that \textit{VideoOrion} outperforms both VideoMAE and UMT-L by more than $2\%$ on average.
\begin{table}[h]
\centering
 \caption{\small Comparison with video-encoder based models.}
 \label{tab:ablate-video-encoder}
 \vspace{-2mm}
 \setlength{\tabcolsep}{2pt}
 \resizebox{0.46\textwidth}{!}{
 \begin{tabular}{lccccc}
 \toprule
 \makecell[l]{Model} & \makecell{MVBench} & \makecell{Egoschema} & \makecell{Perception} & \makecell{VideoMME} & Avg.\\
 \midrule
 VideoMAE & 43.5 & 38.6 & 44.4 & 43.2 & 42.4 \\
 UMT-L & 43.7 & 40.8 & 45.1 & 42.3 & 43.0 \\
 VideoOrion & \textbf{44.2} & \textbf{44.5} & \textbf{46.3} & \textbf{46.1} & \textbf{45.3} \\
 \bottomrule
 \end{tabular}
 }
\vspace{-1mm}
\end{table}

\noindent \textbf{Enhenced Temporal-Understanding.}
The Object-Branch of \textit{VideoOrion} takes in additional frames to capture the essential object dynamic information in the video. We hypothesize that this inclusion of extra temporal information improves the temporal reasoning capabilities.

In Table~\ref{tab:supp-temp}, we compare \textit{VideoOrion} and \textit{VideoOrion$+$} against the baseline models VideoLLaMA2 and VideoLLaMA2.1 on video QA tasks from TemporalBench\citep{cai2024temporalbench}, a benchmark tailored to evaluate fine-grained temporal understanding capabilities. 
The results demonstrate that our models outperform the baselines, underscoring VideoOrion's superior ability to capture and utilize fine-grained temporal details in videos.

\begin{table}[h]
\centering
 \caption{\small Results of zero-shot performance on TemporalBench.}
 \label{tab:supp-temp}
 \scalebox{0.75}{
 \begin{tabular}{lcc}
 \toprule
 \multirow{2}{*}{Model} & Multi-Binary  & Binary  \\
 & Accuracy (Acc.) & Accuracy (Acc.) \\
 \midrule
 VideoLLaMA2 & 15.9 & 57.4  \\
 \textbf{VideoOrion} & \textbf{18.2} & \textbf{59.0}  \\
 \midrule
 VideoLLaMA2.1 & 17.9 & 59.5  \\
 \textbf{VideoOrion$+$} & \textbf{20.3} & \textbf{61.8}  \\

 \bottomrule
 \end{tabular}}
\end{table}

\subsection{Case Study}
\label{sec:case}
This section presents a case study demonstrating how \textit{VideoOrion} utilizes object tokens. Given an input video (Figure~\ref{fig:case}), we ask three different questions and visualize the corresponding changes in attention weights from the last decoder layer of \textit{VideoOrion}'s LLM backbone. The attention weights are normalized so that the sum across all eight object tokens equals one, and we also display the object masks extracted through the detect-segment-track pipeline. The charts illustrate that attention to object tokens varies with different input questions, with higher attention weights assigned to tokens corresponding to more relevant objects.
For instance, the object token for the \textit{person} (last chart) receives the least attention in $Q2$ as it is irrelevant to the question. In $Q1$, attention increases since the question mentions \textit{person}, but it peaks in $Q3$, which focuses solely on the person's features. Conversely, tokens for teabags and cups gain significantly higher attention in $Q2$, directly contributing to the answer.
Taking another perspective, in $Q1$, which concerns the video's general content, attention weights obtained by object tokens are in the middle among the three questions. For the other two questions, which focus on details, attention shifts more toward relevant objects. This suggests that \textit{VideoOrion} adapts its focus dynamically, demonstrating that object tokens effectively enhance video comprehension.
\vspace{-2mm}
\section{Conclusion}
We propose \textit{VideoOrion}, a novel Video-LLM, with explicit disentangled representation for object dynamics in the video. \textit{VideoOrion} has a Video-Centric Branch and an Object-Centric Branch, through a detect-segment-track pipeline with an Object Projector to extract and aggregate the object tokens.  Empirical results across multiple benchmarks demonstrate the capability of \textit{VideoOrion} for improved general video understanding and the inherent ability of referring tasks in videos. 
\vspace{-2mm}
\section*{Limitations}
Despite its effectiveness, our method has certain limitations. The detect-segment-track pipeline relies on multiple vision models, introducing additional computational costs (see Appendix \ref{app:compute} for a detailed analysis) and potentially leading to inaccurate mask extraction, particularly for low-quality videos  (see Appendix \ref{app:fail_pipeline} for a detailed analysis). However, the explicit and disentangled object representation allows the opportunities to diagnose and interpret the mistakes (see Appendix \ref{app:fail_object_attention} for a detailed analysis). Moreover, we believe future advancements in vision models could mitigate these issues, and this work serves as a proof-of-concept that explicit disentangled object representation can enhance general video understanding with inherent referring abilities. Additionally, our framework still relies on the Video-Centric Branch for contextual information, and the alignment between the two branches remains an open area for further investigation.
{
    \small
    \bibliographystyle{ieeenat_fullname}
    \bibliography{main}
}
\clearpage
\setcounter{section}{0}
\renewcommand{\thesection}{\Alph{section}}

\section{Additional Ablation Studies}

Due to space constraint, we provide the rest of the ablation studies here.
\subsection{The Video-Centric Branch}
Given that \textit{VideoOrion} adopts a two-branch design, we investigate the effectiveness of both branches. In Table~\ref{tab:ablate-obj-branch}, we have demonstrated the efficacy of the proposed Object-Centric Branch. Here, in Table~\ref{tab:ablate-video}, we present the performance of the object-only baseline. Relying solely on object tokens results in a performance decline, as only a limited number of tokens—at most 64, or fewer depending on the video—are used to represent the video ($\leq 64$ v.s $576$ for video-branch). This underscores the importance of the video-centric branch in providing contextual information. However, on certain benchmarks (e.g., Perception), the object-only baseline achieves performance comparable to the video-only baseline (with 576 tokens for the video), suggesting that object tokens capture essential information.

\begin{table}[h]
\centering
 \caption{\small Ablation study on the Video-Centric Branch.}
 \label{tab:ablate-video}
 \vspace{-2mm}
 \setlength{\tabcolsep}{2pt}
 \resizebox{0.46\textwidth}{!}{
 \begin{tabular}{lccccc}
 \toprule
 \makecell[l]{Model} & \makecell{MVBench} & \makecell{Egoschema} & \makecell{Perception} & \makecell{VideoMME} & Avg.\\
 \midrule
 object-only& 37.0 & 34.3 & 43.3 & 40.8 & 38.9 \\
 VideoOrion & \textbf{44.2} & \textbf{44.5} & \textbf{46.3} & \textbf{46.1} & \textbf{45.3} \\
 \bottomrule
 \end{tabular}
 }
\vspace{-1mm}
\end{table}

\subsection{Number of Object Tokens}
We also study the impact of different upper limits for the number of object tokens $N_{o_i}$ in Table~\ref{tab:ablate-obj-num}.

\begin{table}[h]
\centering
 \caption{\small Different upper limit numbers of object tokens.}
 \label{tab:ablate-obj-num}
 \vspace{-2mm}
 \setlength{\tabcolsep}{2pt}
 \resizebox{0.46\textwidth}{!}{
 \begin{tabular}{cccccc}
 \toprule
 \makecell[l]{Max $N_{o_i}$} & \makecell{MVBench} & \makecell{Egoschema} & \makecell{Perception} & \makecell{VideoMME} & Avg.\\
 \midrule
 16 & 43.7 & 41.1 & 44.7 & 45.3 & 43.7 \\
 32 & 43.9 & 40.9 & 45.8 & 45.2 & 44.0 \\
 \rowcolor{gray!20} 64 & \textbf{44.2} & \textbf{44.5} & \textbf{46.3} & \textbf{46.1} & \textbf{45.3} \\
 80 & 44.0 & 40.2 & 45.4 & 45.3 & 43.7 \\
 \bottomrule
 \end{tabular}
 }
\vspace{-1mm}
\end{table}



\section{More Examples of the Detect-Segment-Track Pipeline}
\begin{figure}[h]
    \centering
    \includegraphics[width=0.47\textwidth]{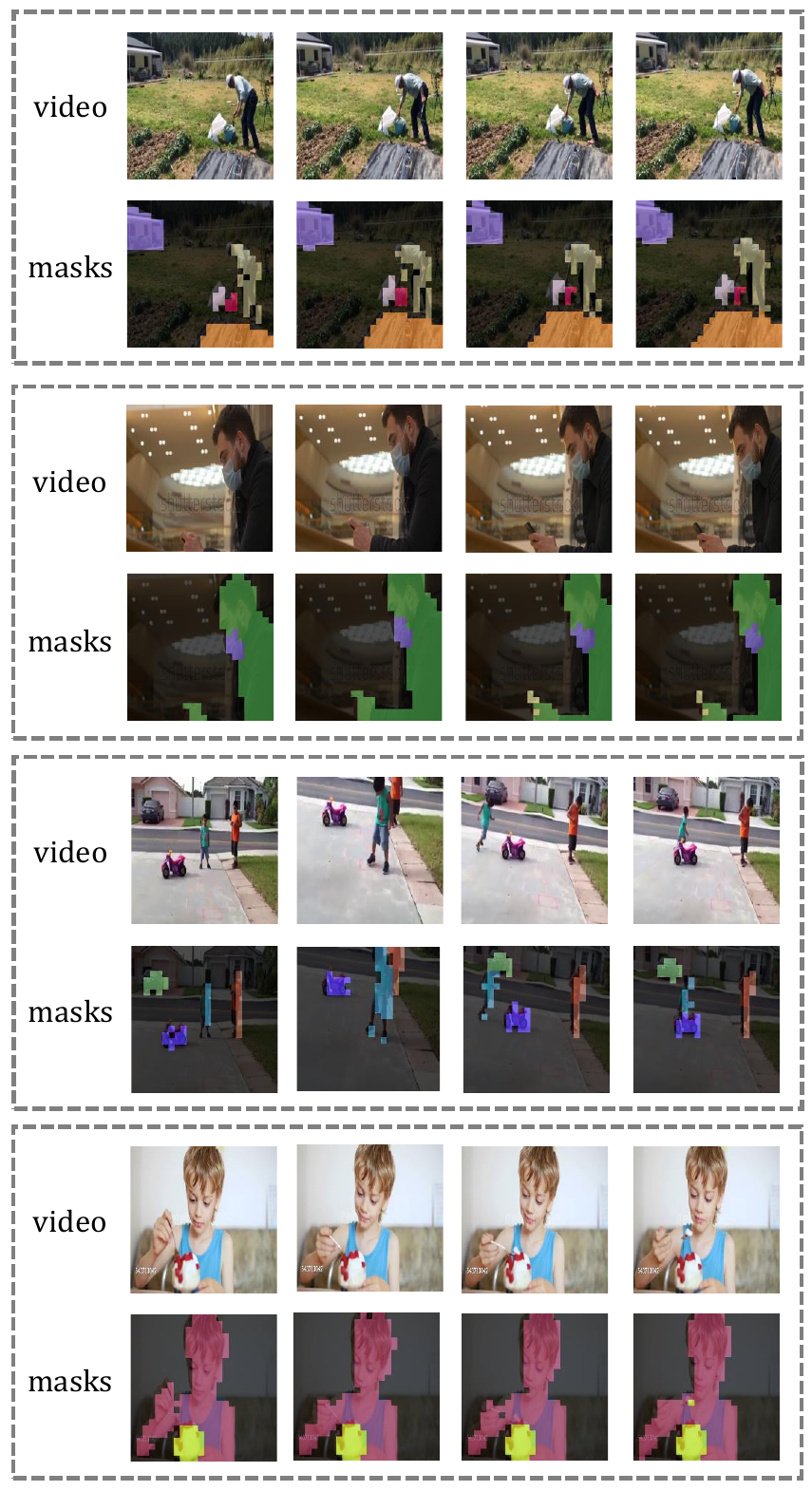}
    \caption{Examples of the detect-segment-track pipeline.}
    \label{fig:supp-detect}
\end{figure}

We show additional examples of the object mask lists extracted through the detect-segment-track pipeline in Figure~\ref{fig:supp-detect}. 
To povide a clearer illustration of the mask pooling mechanism in our model, we resize the masks and map them to the patch level.
As can be seen in most instances, the pipeline effectively identifies the salient objects present in videos, ensuring that the resulting object tokens are enriched with clear and meaningful semantics.

\section{More Qualitative Results}

\begin{figure}[h]
    \centering
    \includegraphics[width=0.47\textwidth]{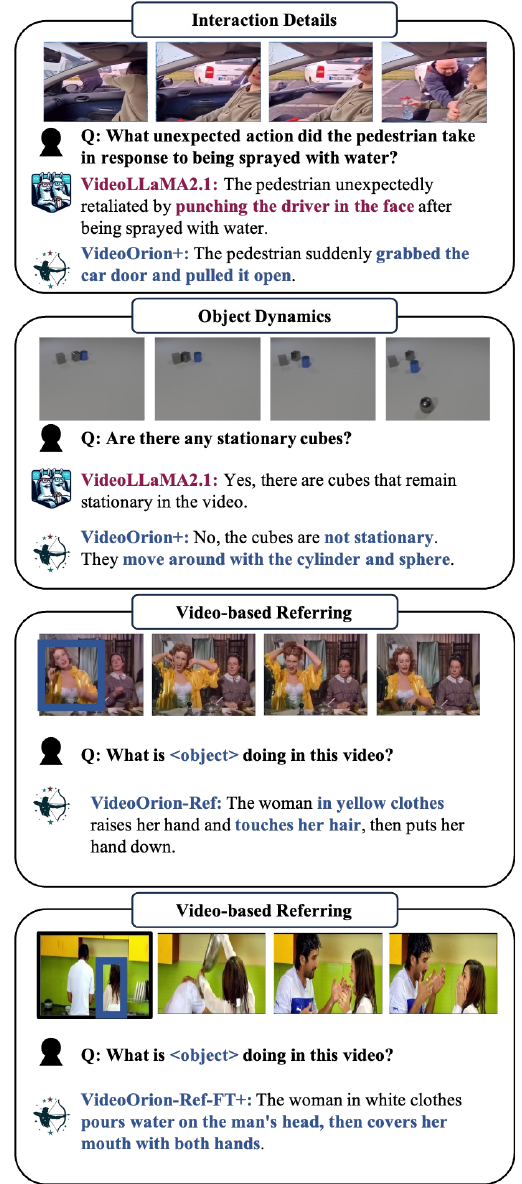}
    \caption{Qualitative examples of \textit{VideoOrion$+$}, \textit{VideoOrion-Ref} and \textit{VideoOrion-Ref-FT$+$}.}
    \label{fig:supp-case}
\end{figure}

Additional qualitative examples of \textit{VideoOrion$+$}, \textit{VideoOrion-Ref} and \textit{VideoOrion-Ref-FT$+$}  are presented in Figure~\ref{fig:supp-case}. 
These examples highlight our model's capabilities of capturing interaction details and object dynamics, as well as its enhanced video-based referring capabilities after being trained on this task.

\section{Analysis of Failure Cases}
\subsection{Failures in Detect-Segment-Track Pipeline}
\label{app:fail_pipeline}
\begin{figure}[h]
    \centering
    \includegraphics[width=0.47\textwidth]{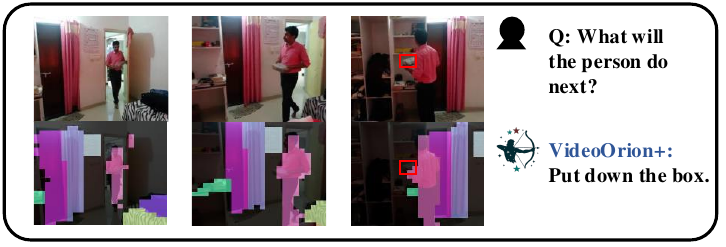}
    \caption{Failure of the detect-segment-track pipeline.}
    \label{fig:supp-fail}
\end{figure}

One potential limitation is that inaccuracies in the pipeline may hinder the model's understanding and perception abilities. However, the dual-branch design of  \textit{VideoOrion} helps alleviate this issue by leveraging context tokens as complementary. As shown in Figure~\ref{fig:supp-fail}, even when the pipeline fails to detect and track the box in a person's hand,  \textit{VideoOrion} can still correctly infer the action based on contextual cues.

\subsection{Failures on Object Attention}
\label{app:fail_object_attention}
Although \textit{VideoOrion} successfully detects and tracks critical objects, it can still occasionally make errors. However, its explicit and disentangled object representation allows for better diagnosis and interpretation of these mistakes.

\begin{figure}[h]
    \centering
    \includegraphics[width=0.47\textwidth]{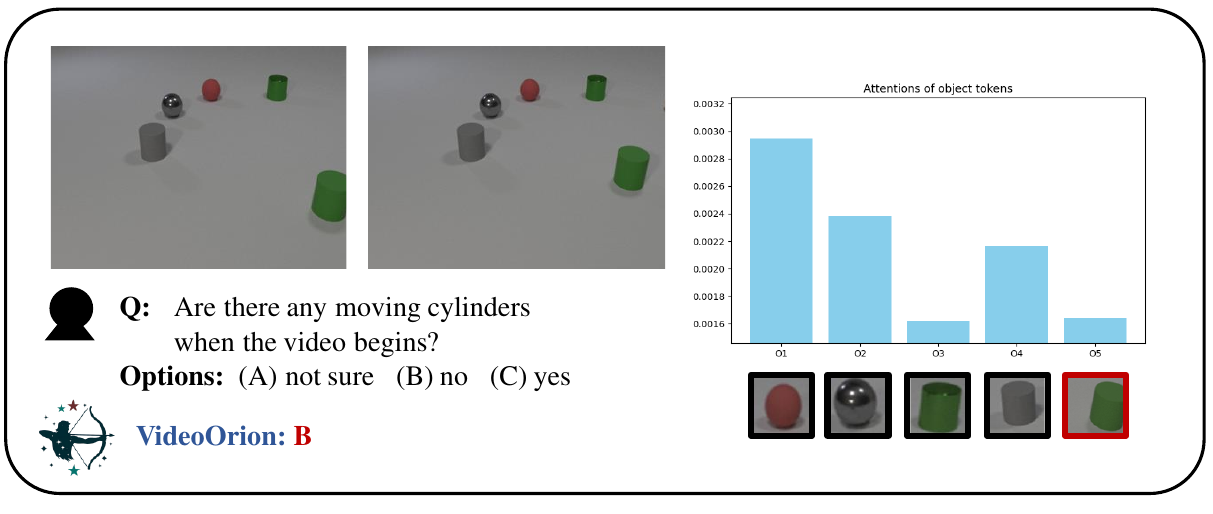}
    \caption{Failure case due to the object attention.}
    \label{fig:supp-fail-attn}
\end{figure}

We analyze a case presented in Figure~\ref{fig:supp-fail-attn}, where a green cylinder in the bottom right corner moves, yet \textit{VideoOrion} incorrectly predicts that no cylinders are moving. By visualizing the attention weights assigned to the object tokens, we observe that the model assigns relatively low attention to the moving cylinder ($O5$) while focusing more on the static grey cylinder ($O4$). This likely explains the misclassification in this instance. This case highlights that while object tokens generally enhance \textit{VideoOrion}'s understanding, misplaced attention on irrelevant objects can still lead to errors.

\section{Additional Computation Time Induced}
\label{app:compute}
As noted in Limitations, our method will increase computation cost due to the additional Detect-Segment-Track pipeline. We report an average computation time for 50 samples in Table~\ref{tab:supp-compute} and observe a 38.5\% increase. We believe this computation cost is acceptable in trade-off of the benefits brought by \textit{VideoOrion}. We also hypothesize that with computation optimization and faster tracking model, these extra time will be negliable in the future.

\begin{table*}
\centering
 \caption{\small Additional computation time for \textit{VideoOrion}.}
 \label{tab:supp-compute}
\begin{tabular}{llll}
 \toprule
  & no pipeline & with pipeline & extra (\%) \\
 \midrule
 Time/sample & $8.27$s & $11.46$s & $+38.5\%$\\
 \bottomrule
\end{tabular}
\end{table*}

\section{Scaling Effect Observed in \textit{VideoOrion}}
\begin{figure}[h]
    \centering
    \includegraphics[scale=0.28]{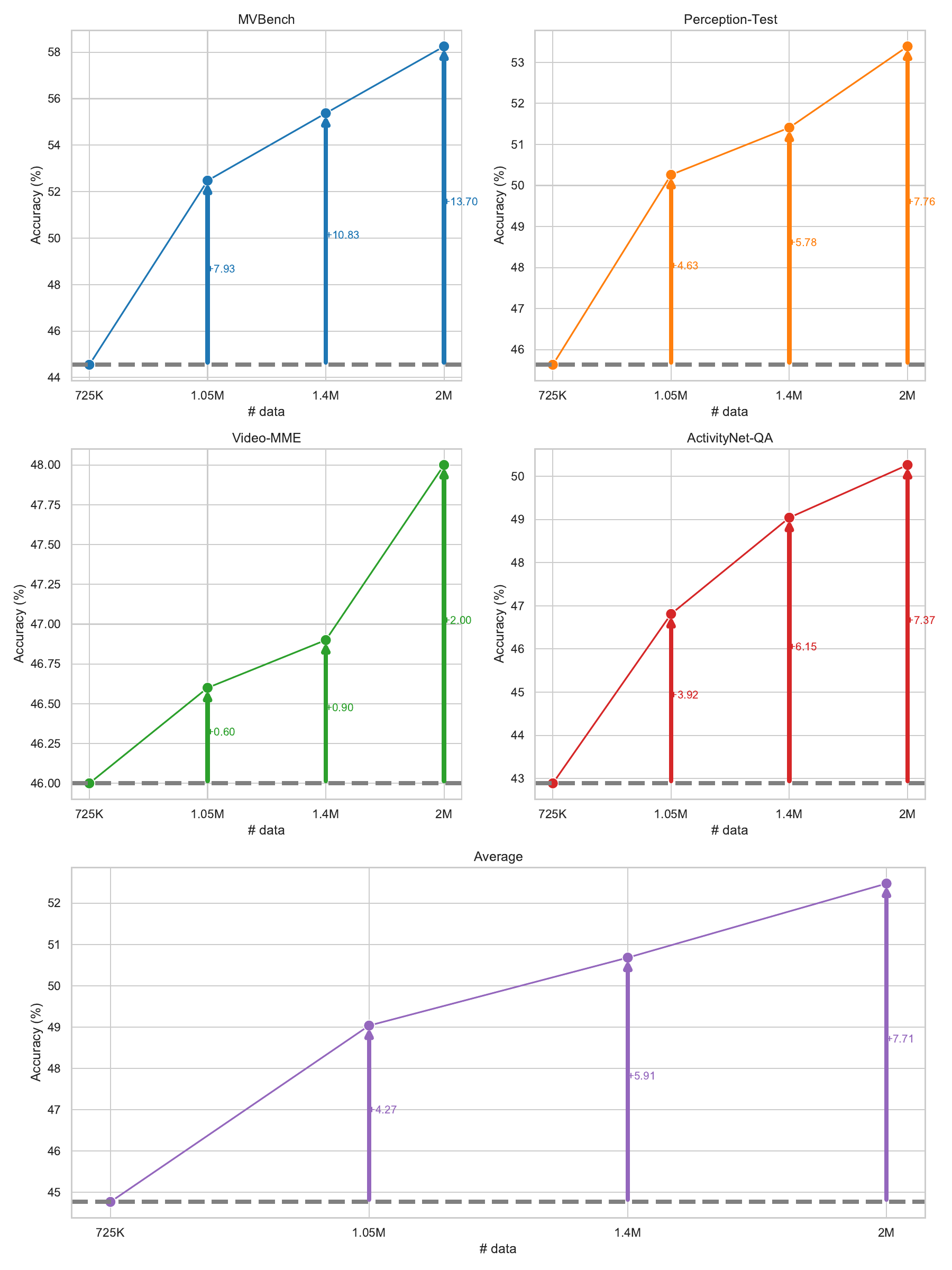}
    \caption{An illustration showcasting how \textit{VideoOrion} benefits from increased training data.}
    \label{fig:supp-data}
\end{figure}

In this section, we demonstrate how our model can benefit from data scaling. 
To evaluate this, we randomly 
divide the instruction tuning dataset from VideoChat2 into three parts, and we begin by fine-tuning \textit{VideoOrion} using only the Video-LLaVA dataset. Subsequently, we progressively incorporate each of the three parts from VideoChat2 into the training data and demonstrate how performance evolve with scaling of the dataset.
As per Figure~\ref{fig:supp-data},
the performances of \textit{VideoOrion} consistently improve across all four benchmarks, i.e. MVBench, Perception-Test, Video-MME and ActivityNet-QA, with more training data.
These results demonstrate \textit{VideoOrion}'s capacity to effectively harness larger datasets, enabling consistent improvements in performance.

\section{Hyperparameters}
We report in Table~\ref{tab:supp-hyper} the detailed hyperparameters for \textit{VideoOrion} and \textit{VideoOrion$+$} used in different training stages.
\begin{table*}
\centering
 \caption{\small Training hyperparameters for \textit{VideoOrion} and \textit{VideoOrion$+$}.}
 \label{tab:supp-hyper}
 \setlength{\tabcolsep}{8pt}
 \scalebox{1}{
 \begin{tabular}{l|ccc|ccc}
 \toprule
 \multirow{2}{*}{Config} & \multicolumn{3}{c|}{\textbf{VideoOrion}} & \multicolumn{3}{c}{\textbf{VideoOrion$+$}} \\
 & Stage1 & Stage2 & Stage3 & Stage1 & Stage2 & Stage3 \\
 \midrule

 Vision Encoder & \multicolumn{3}{c|}{CLIP(ViT-L/14)} & \multicolumn{3}{c}{SigLIP(so400m-patch14-384)} \\
 LLM Backbone & \multicolumn{3}{c|}{Mistral-Instruct-7B} & \multicolumn{3}{c}{Qwen2-7B} \\
 Frame Number & 8 & 8 & 8 & 16 & 16 & 16\\
 Input Resolution & 336 & 336 & 336 & 384 & 384 & 384 \\
 Learning Rate & $1e-3$ & $1e-4$ & $5e-6$ & $1e-3$ & $1e-4$ & $5e-6$ \\
 Weight Decay & 0 & 0 & 0 & 0 & 0 & 0 \\
 Warmup Ratio & $0.03$ & $0.03$ & $0.03$ & $0.03$ & $0.03$ & $0.03$ \\
 Learning Rate Schedule & cosine & cosine & cosine & cosine & cosine & cosine \\
 Numerical Precision & bfloat16 & bfloat16 & bfloat16 & bfloat16 & bfloat16 & bfloat16 \\
 Batch Size & 256 & 256 & 128 & 256 & 256 & 128 \\
 LLM Sequence Length & 2048 & 2048 & 2048 & 2048 & 2048 & 2048 \\
 Epoch Number & 1 & 1 & 1 & 1 & 1 & 1 \\
 Max Object Token Number & - & - & 64 & - & - & 64 \\

 \bottomrule
 \end{tabular}}
\end{table*}




\end{document}